\documentclass[10pt,twocolumn,letterpaper]{article}

\usepackage{cvpr}
\usepackage{times}
\usepackage{epsfig}
\usepackage{graphicx}
\usepackage{amsmath}
\usepackage{amssymb}

\usepackage{booktabs}
\usepackage{multirow,tabularx}
\usepackage{enumitem}
\setlist{nosep}
\usepackage{tabularx}
\usepackage{soul}
\usepackage{balance}
\usepackage{subfigure}
\usepackage{booktabs}
\usepackage[multiple]{footmisc}
\usepackage{xcolor}
\usepackage{authblk}
\usepackage{colortbl}
\definecolor{lightgray}{gray}{0.9}

\usepackage{float}
\usepackage{stfloats}

\usepackage[breaklinks=true,bookmarks=false]{hyperref}

\cvprfinalcopy 

\def\cvprPaperID{3} 

\ifcvprfinal\fi
\begin{document}

\title{Fashion-Guided Adversarial Attack on Person Segmentation}

\author[1]{Marc Treu\thanks{Equal contributions. This work was done when Marc Treu interned at NII, Japan. Contact email: \href{mailto:marc.treu@tehtris.com}{marc.treu@tehtris.com}.}}
\author[2]{Trung-Nghia Le$^*$}
\author[2]{Huy H. Nguyen$^*$}
\author[2,3]{Junichi Yamagishi}
\author[2,3,4]{Isao Echizen}

\affil[1]{TEHTRIS, France}
\affil[2]{National Institute of Informatics, Japan}
\affil[3]{The Graduate University for Advanced Studies, SOKENDAI, Japan}
\affil[4]{University of Tokyo, Japan}

\maketitle

\begin{abstract}
This paper presents the first adversarial example based method for attacking human instance segmentation networks, namely person segmentation networks in short, which are harder to fool than classification networks. We propose a novel Fashion-Guided Adversarial Attack (FashionAdv) framework to automatically identify attackable regions in the target image to minimize the effect on image quality. It generates adversarial textures learned from fashion style images and then overlays them on the clothing regions in the original image to make all persons in the image invisible to person segmentation networks. The synthesized adversarial textures are inconspicuous and appear natural to the human eye. The effectiveness of the proposed method is enhanced by robustness training and by jointly attacking multiple components of the target network. Extensive experiments demonstrated the effectiveness of FashionAdv in terms of robustness to image manipulations and storage in cyberspace as well as appearing natural to the human eye. The code and data are publicly released on our project page\footnote{\url{https://github.com/nii-yamagishilab/fashion\_adv}}.
\end{abstract}

\section{Introduction}
\label{sec:introduction}

Powerful computer vision methods have been applied to object classification~\cite{He-CVPR2016, Huang-CVPR2017}, object detection~\cite{ltnghia-WACV2020, Ren-NIPS2015}, semantic segmentation~\cite{ltnghia-CVIU2019, Long-ICCV2015}, and instance segmentation~\cite{Kaiming-ICCV2017, ltnghia-WACV2019}, enabling machines to perceive the world. With incremental learning using a massive amount of data obtained from the Internet and surveillance cameras, vision-based intelligent systems (\textit{i.e.}, face recognition~\cite{Sobel-2020-SSRN}, and gait recognition~\cite{Wan-2018-ACM}) can track and learn the behaviors of a large population. Such systems are typically trained using images and videos crawled from social networks without authorization, which raises a serious privacy issue requiring governments and companies to establish policies to prevent unauthorized surveillance and tracking on the basis of user data~\cite{Sobel-2020-SSRN}. Moreover, social network users need a tool to protect their privacy. The tool should be robust against the transformations and compressions that occur during the uploading, sharing, and storing images and videos.

\begin{figure}[t!]
	\centering
	\includegraphics[width=1\linewidth]{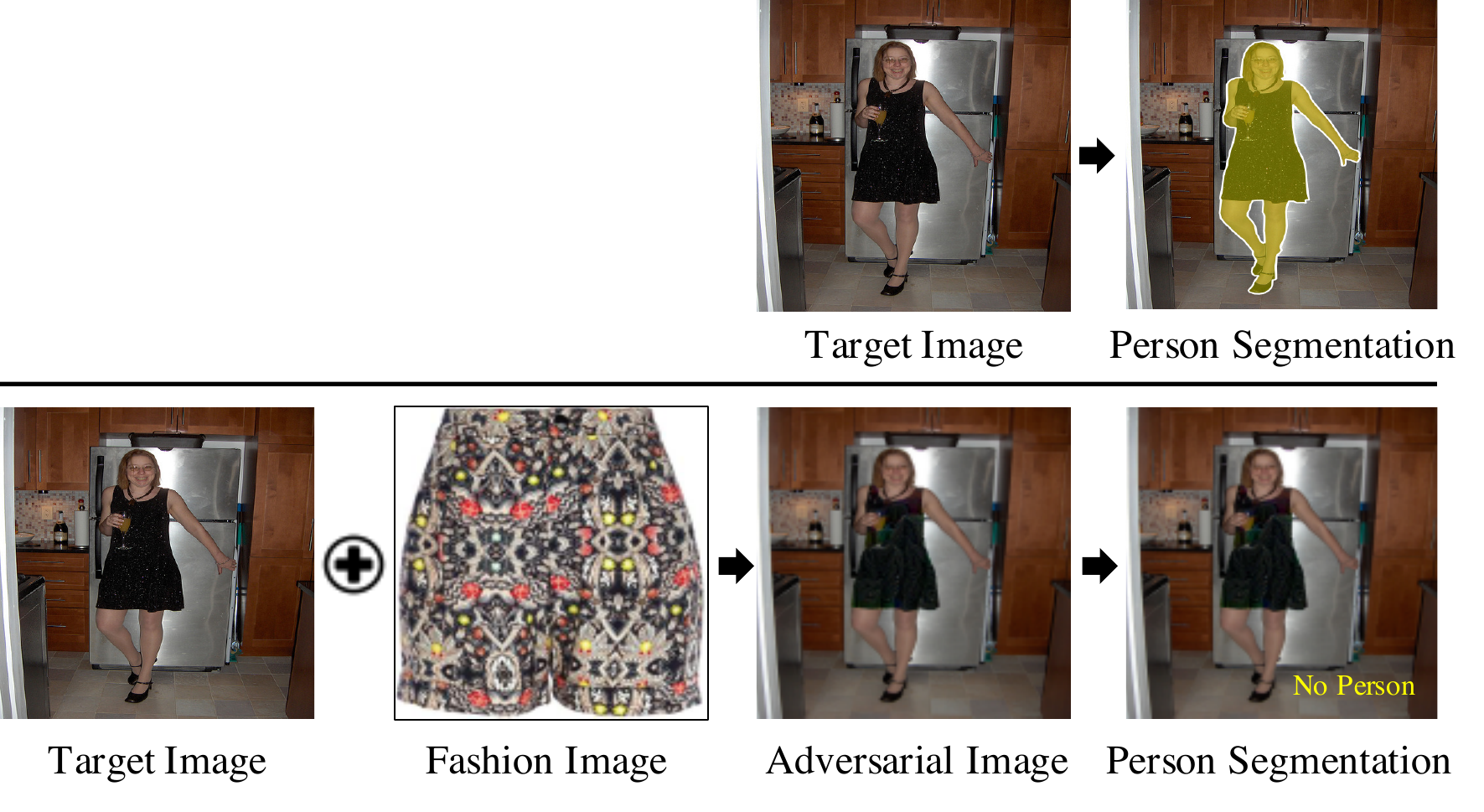}
	\caption{Overview of our proposed Fashion-Guided Adversarial Attack (FashionAdv). Top row: person is segmented successfully. Bottom row: given a target image and a guided-fashion style image, FashionAdv synthesizes natural clothing texture that can nevertheless fool a person segmentation network by making the person invisible to the network.}
	\label{fig:overview}
\end{figure}

\begin{figure}[t!]
	\centering
	\includegraphics[width=1\linewidth]{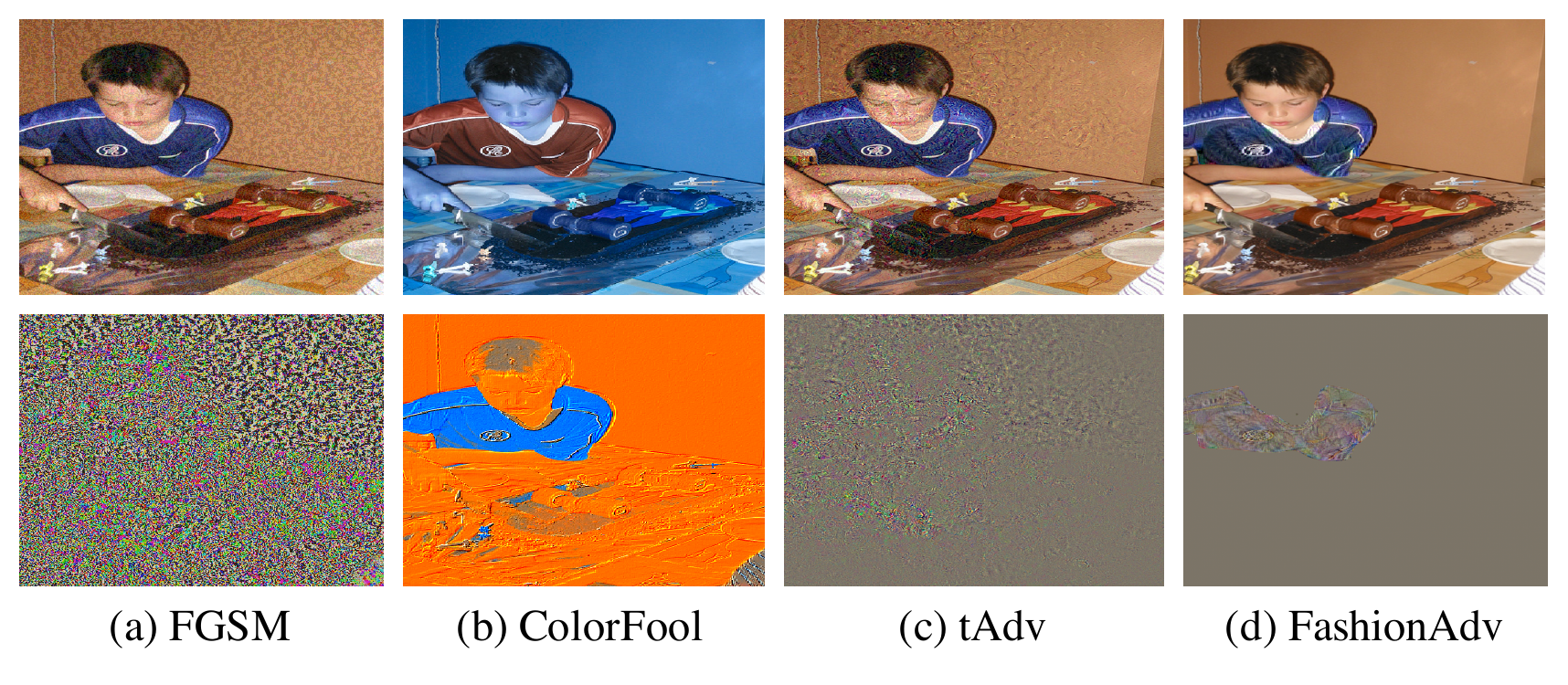}
	\caption{Effectiveness of our proposed FashionAdv, compared with conventional digital-level adversarial attacks. From left to right: adversarial noise attack~\cite{Goodfellow-2014-ICLR}, adversarial color attack~\cite{Shamsabadi-2020-CVPR}, adversarial texture attack~\cite{Bhattad-ICLR2020}, and our FashionAdv. Bottom row: corresponding adversarial perturbations. Changes made by FashionAdv are inconspicuous to the human eye.}
	\label{fig:examples}
	\vspace{-3mm}
\end{figure}

Several attack methods using adversarial examples have been developed to degrade the performance of deep neural networks in the digital world. Most such methods effectively fool object classification by directly attacking the digital image or video files~\cite{Bhattad-ICLR2020, Shamsabadi-2020-CVPR}. Adversarial noise-based attack methods were developed first and became well-established~\cite{Goodfellow-2014-ICLR, Kurakin-2016-ICLRW, Madry-ICLR2018, Moosavi-CVPR2016}). Subsequently, attacks based on color~\cite{Shamsabadi-2020-CVPR} and texture~\cite{Bhattad-ICLR2020, Duan-CVPR2020}, which also target human perception, have been developed. A few attack methods have been developed to target object detection and semantic segmentation~\cite{Xie-ICCV2017}. To the best of our knowledge, a method for attacking instance segmentation, which is a combination of object detection and semantic segmentation, has not been developed. In this work, we develop a method for targeted adversarial attacks against human instance segmentation, namely person segmentation. 

Naturalness and robustness are two essential properties of adversarial examples~\cite{Bhattad-ICLR2020, Zhao-2019-SIGSAC}. Naturalness means that adversarial examples are unlikely to be noticed by people, while robustness means that they are not paralyzed by data compression, transformation, or image filter effects (\eg, beauty applications). There is a trade-off between these properties. Adversarial examples created using traditional approaches can deal with naturalness~\cite{Bhattad-ICLR2020, Shamsabadi-2020-CVPR}, but they are easily negated by image transformation~\cite{Guo-2018-ICLR} or compression~\cite{Goodfellow-2014-ICLR}. \textit{This work aims to balance this trade-off}.

To this end, we propose an attack method for use against person segmentation network (\ie, YOLACT~\cite{Bolya-ICCV2019}, one of the first networks attempting real-time instance segmentation). Our proposed Fashion-Guided Adversarial Attack (FashionAdv) is designed to synthesize \textit{\textbf{adversarial textures} by blending generated adversarial noise transferred from customizable fashion style images into the original clothing regions}. The new image with adversarial textures, an adversarial image, is used to spoof the target person segmentation network by making the persons invisible to the network (\cf Fig.~\ref{fig:overview}). 
Directly optimizing the adversarial texture through infusion of texture from the fashion style image helps make the synthesized adversarial texture inconspicuous and appear natural to the human eye despite their large perturbations. Indeed, a quick visual comparison of the results with FashionAdv with those of conventional adversarial attacks in Fig.~\ref{fig:examples} indicates that FashionAdv is capable of generating highly inconspicuous adversarial textures. Unlike conventional adversarial attacks~\cite{Bhattad-ICLR2020, Goodfellow-2014-ICLR, Shamsabadi-2020-CVPR}, which attack the entire image, FashionAdv automatically detects the attackable regions (\ie, clothing regions) to minimize the impact of adversarial examples on image quality. Only clothing regions are identified as attackable regions because (1) these regions are suitable for embedding adversarial noise while maintaining image quality, and (2) human perception is less sensitive to these regions compared with other regions such as faces~\cite{Leopold-JCP2010}. FashionAdv effectively works on all persons in the image, hiding them from deep instance segmentation networks. Moreover, we propose jointly attacking two network components, the classification and the segmentation sub-networks. Multiple cues are integrated into the loss function, making the adversarial attack robust against image manipulations and storage in cyberspace. 

\begin{figure}[t!]
	\centering
	\vspace{-3mm}
	\includegraphics[width=1\linewidth]{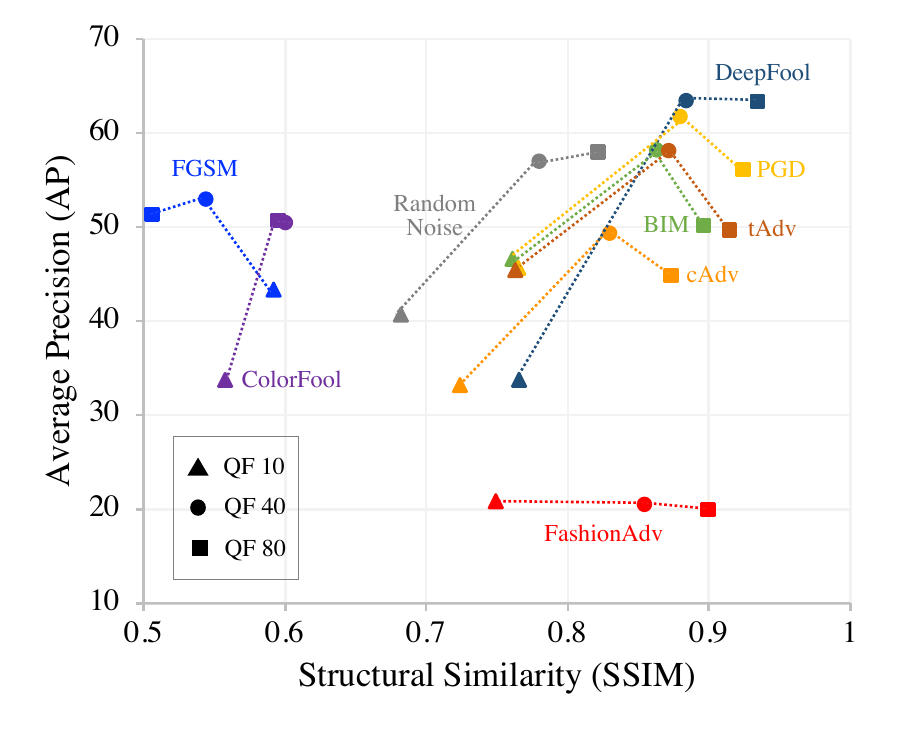}
	\caption{Naturalness (\ie, SSIM~\cite{Wang-TIP2004}) v.s. robustness (\ie, AP~\cite{Lin-ECCV2014}) on MS-COCO dataset. FashionAdv significantly outperformed conventional adversarial example methods in terms of robustness against JPEG compression (quality factors 10, 40, and, 80) while maintaining naturalness competitive with that of the conventional methods. Hence, FashionAdv successfully spoofed the target person segmentation network (\ie, YOLACT~\cite{Bolya-ICCV2019}), and the changes made were inconspicuous to the human eye.}
	\label{fig:trade_off}
	\vspace{-3mm}
\end{figure}

Extensive experiments on the MS-COCO dataset~\cite{Lin-ECCV2014} demonstrated that the \textit{changes made by FashionAdv are not only robust against image manipulations and storage in cyberspace} (\eg, image filters and JPEG compression) \textit{but are also inconspicuous and appear natural to the human eye despite having large perturbations} (\cf Fig.~\ref{fig:trade_off}). The code and data are made available on our project page\footnote{\url{https://github.com/nii-yamagishilab/fashion\_adv}}.

Our contributions are as follows. 

\begin{itemize}
	\item We have developed a novel adversarial attack method for use against person segmentation networks that achieves both robustness and naturalness. Our Fashion-Guided Adversarial Attack (FashionAdv) is the first reported attack method for use against instance segmentation.
	\item FashionAdv flexibly synthesizes clothing textures from guided-fashion style images that are inconspicuous and appear natural to the human eye despite their large perturbations. 
	\item FashionAdv focuses on only the clothing regions in images, which minimizes the effect on image quality. It is also robust against image manipulations and storage in cyberspace, such as image filtering and JPEG compression, respectively.
	\item Extensive experiments demonstrated that FashionAdv significantly outperforms conventional methods in term of robustness while still maintaining naturalness.
\end{itemize}

\section{Related Work}
\label{sec:related_work}

\subsection{Adversarial Attacks}

There are two types of adversarial attacks, including digital-level and physical-level. The former target object classification, while the latter target both object classification and object detection. In digital-level attacks, adversarial perturbations are added to digital images or videos~\cite{Akhtar-Access2018}. In physical-level attacks, real adversarial objects are crafted~\cite{Kurakin-2016-ICLRW}. For both levels, the adversarial perturbations can be noise~\cite{Luo-2018-AAAI}, patterns~\cite{Duan-CVPR2020, Huang-2020-CVPR, Zhang-2019-ICLR}, colors~\cite{Shamsabadi-2020-CVPR, Bhattad-ICLR2020}, or patches~\cite{Brown-2017-NIPSW, Thys-2019-CVPRW, Zhao-2019-SIGSAC}.

Most \textbf{digital-level adversarial attacks} target object classification~\cite{Akhtar-Access2018}. Adversarial noise-based attacks (\eg, gradient-based attacks) were developed first and became well-known, including FGSM~\cite{Goodfellow-2014-ICLR}, BIM~\cite{Kurakin-2016-ICLRW}, DeepFool~\cite{Moosavi-CVPR2016}, and PGD~\cite{Madry-ICLR2018}). FGSM~\cite{Goodfellow-2014-ICLR} uses gradients of the target neural network to craft adversarial examples. BIM~\cite{Kurakin-2016-ICLRW} performs multiple FGSM attacks with small step size. DeepFool~\cite{Moosavi-CVPR2016} estimates the minimal adversarial perturbation by using a simple iterative method. PGD~\cite{Madry-ICLR2018} treats adversarial attacks as a constrained optimization problem.

Furthermore, attacks based on color and texture (\eg, unrestricted attacks), which also target human perception, have been developed. Shamsabadi \etal~\cite{Shamsabadi-2020-CVPR} devised ColorFool attack in which colors within a certain range are modified rather than adversarial noise or patches being crafted. Duan \etal~\cite{Duan-CVPR2020} used both colors and patterns to generate photorealistic adversarial examples. Bhattad \etal~\cite{Bhattad-ICLR2020} introduced two attacks that use realistic color perturbation (cAdv) and texture extracted from a source image (tAdv). 

Most \textbf{physical adversarial attacks} target object detection. The Expectation Over Transformation (EOT) framework~\cite{Athalye-2018-ICML} was originally developed for object classification, and then was extended to attack object detection. Thys \etal~\cite{Thys-2019-CVPRW} used adversarial patches to attack a human detection system using YOLO-v2~\cite{Redmon-CVPR2017}. Zhao \textit{et-al.}~\cite{Zhao-2019-SIGSAC} proposed two types of adversarial patches (\ie, hiding attack and appearing attack) to attack Faster RCNN~\cite{Ren-NIPS2015} and YOLO-v3~\cite{Redmon-2018}. Zhang \etal~\cite{Zhang-2019-ICLR} learned camouflaged adversarial patterns to attack Faster RCNN~\cite{Ren-NIPS2015} and YOLO~\cite{Redmon-CVPR2016}. Following this, Huang \etal~\cite{Huang-2020-CVPR} made camouflage adversarial patterns universal. On the other hand, few adversarial patch-based methods were proposed for attacking object recognition in the real-world, especially road objects, which is important for autonomous driving systems~\cite{Brown-2017-NIPSW, Duan-CVPR2020, Luo-2018-AAAI}. In addition, few methods have been recently developed to attack person detectors in the real-world by printing adversarial patches on t-shirts~\cite{Wu-ECCV2020, Kaidi-ECCV2020}. However, these methods look extremely unnatural to the human eye and adversarial patches must be manually attached to other objects (\ie~t-shirts).

In this work, \textit{we focus on attacks based on adversarial examples in cyberspace (digital level)}. According to the best of our knowledge, there has been no work on attacking instance segmentation. Therefore, in this work, we target adversarial attacks for use against person segmentation. Particularly, we target YOLACT~\cite{Bolya-ICCV2019}, one of the first networks attempting real-time instance segmentation. YOLACT breaks instance segmentation into two parallel subtasks (\ie, generating a set of prototype masks and predicting per-instance mask coefficients) and then linearly combines the prototypes with the mask coefficients. We note that we target only persons segmented from YOLACT.

\subsection{Improvements on Robustness and Naturalness}

Easily created adversarial examples can fool several kinds of deep neural networks, including ones for object detection and semantic segmentation~\cite{Akhtar-Access2018}. However, adversarial perturbations created using traditional approaches can be negated by image transformations~\cite{Guo-2018-ICLR, Luo-2015-arXiv}, or compression~\cite{Goodfellow-2014-ICLR}. Recent work has been carried out to improve the robustness of adversarial perturbations, especially in the physical world. Physical adversarial objects crafted based on the basis of the EOT framework~\cite{Athalye-2018-ICML} are robust to a combination of noise, distortion, and affine transformation. The algorithm has thus been integrated into different attacks to improve robustness~\cite{Brown-2017-NIPSW, Qin-2019-ICML, Wiyatno-2019-ICCV}.

A comprehensive benchmark recently performed by Dong  \etal~\cite{Dong-2020-CVPR-Benchmarking} provides an overview of the robustness of several adversarial attacks. By considering the limitations of pixel-wise and global adversarial attack methods, Dong \etal~\cite{Dong-2020-CVPR-Superpixed} developed an attention-based attack that focuses only on salient objects to improve robustness. Luo \etal~\cite{Luo-2018-AAAI} introduced a human-perception-based metric for estimating the distance between pixels and a greedy optimization algorithm to maximize the noise tolerance of adversarial examples. Zhao \etal~\cite{Zhao-2019-SIGSAC} used feature-interference reinforcement and enhanced realistic constraints generation to enhance the robustness of their proposed hiding attack.

Although naturalness is essential for human perception, this property has not been well explored in the literature. There has been little work dealing with the naturalness of adversarial attacks. Shamsabadi \etal~\cite{Shamsabadi-2020-CVPR} simply modified colors in the target image within ranges that are perceived as natural by humans. Bhattad \etal~\cite{Bhattad-ICLR2020} searched texture sources in feature space to find the one most similar to the target image and used it to generate more natural-looking images. Duan \etal~\cite{Duan-CVPR2020} transferred adversarial perturbations into natural-looking styles, namely camouflage styles that appear natural to human observers. Wu \etal\cite{Wu-ECCV2020} trained adversarial patterns from real images and printed them on posters and T-shirts for physical-level attacks.


\begin{figure*}[t!]
	\centering
	\includegraphics[width=1\textwidth]{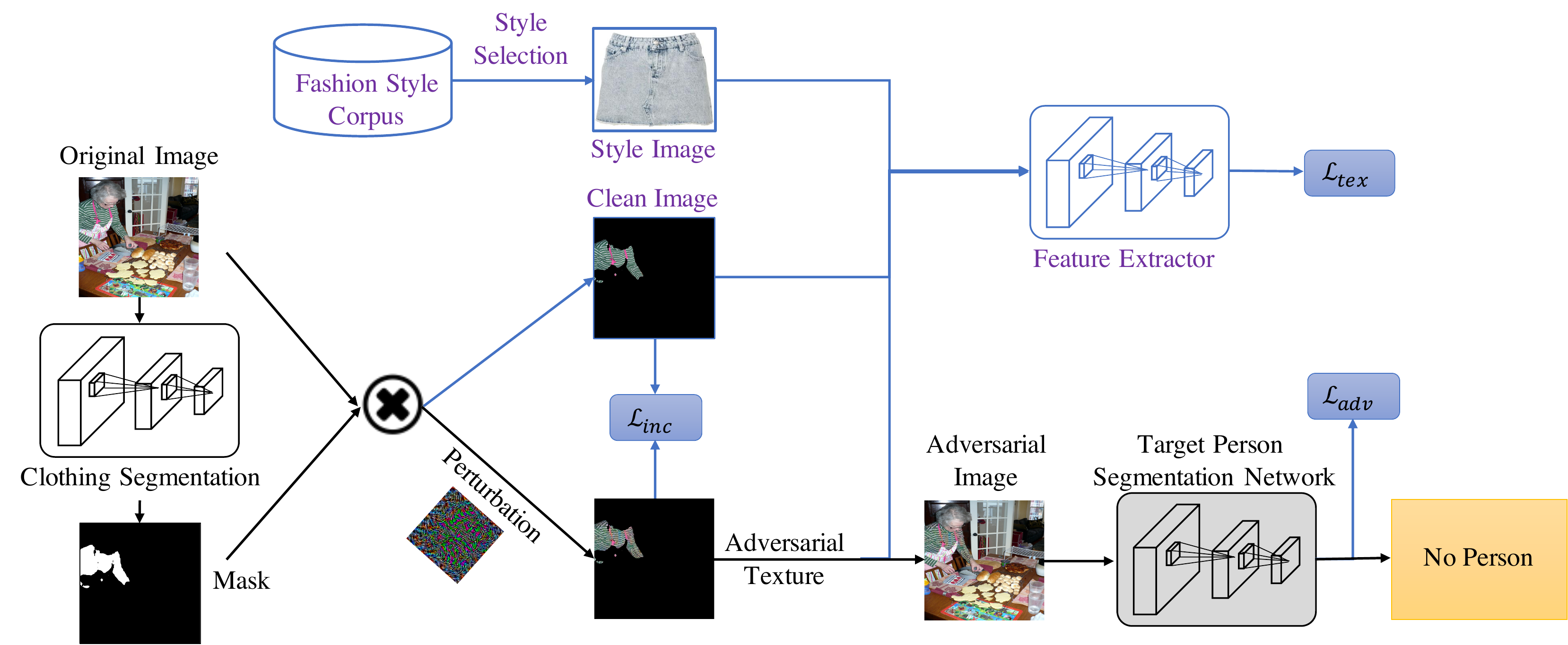}
	\caption{Workflow of proposed FashionAdv framework. Given a target image, our FashionAdv identifies attackable regions in the image and generates an adversarial texture from a guided-fashion style image, which is the image closest to the target image automatically obtained by searching our constructed fashion corpus. The optimized adversarial image is used to spoof a person segmentation network. Blue flow (\ie, blue arrows and blue blocks) illustrates the learning process only.}
	\label{fig:FashionAdv}
\end{figure*}

\section{Proposed Method}
\label{sec:proposed_method}

\subsection{Overview}

We overview our proposed FashionAdv method by describing how it can be used to attack a state-of-the-art instance segmentation network (\ie, YOLACT~\cite{Bolya-ICCV2019}), particularly targeting person segmentation (\ie, human instance segmentation). By generating adversarial textures and painting them on clothing regions, FashionAdv should cause YOLACT to fail in segmenting individual persons under diverse perturbation transformation conditions (\ie, image compression, image blurring, noise addition, color variation, and perspective transformation).

Given an image $X$ of size $(w, h)$ consisting of $K$ persons, FashionAdv decomposes the image into person regions $P$ and non-person regions $\bar{P}$. It focuses on only the person regions:
\begin{equation}
P = \{P_k: P_k = X \cdot M_k + X \cdot \bar{M}_k\}_{k=1}^K,
\end{equation}
where $P_k$ indicates the $k$-th person, and $M_k,\bar{M}_k\in\{0,1\}^{w,h}$ are binary masks of the attackable regions (\eg, clothing) and unattackable regions (\eg, face, hair, hand, and leg) of person $P_k$, respectively. These binary masks specify the locations to which the pixels of image $X$ belong, and "$\cdot$" denotes pixel-wise multiplication.

FashionAdv manipulates only the attackable regions (\eg, clothing), corresponding to masks $M_k$ instead of the entire image, resulting in an adversarial image $\tilde{X}$. That is, it creates an adversarial texture for the attackable regions of each person $\tilde{P}_k$: 
\begin{equation}
\tilde{P} = \{\tilde{P}_k: \tilde{P}_k = \tilde{X} \cdot M_k + X \cdot \bar{M}_k\}_{k=1}^K.
\end{equation}

Our goal is to generate adversarial image $\tilde{X}$ with both robustness and naturalness. For \textit{naturalness}, FashionAdv synthesizes adversarial textures $\tilde{P}_k$ from fashion style image $S$ and minimizes the distance between $\tilde{P}_K$ with both $P_K$ and $S$. This learning criterion should yield adversarial textures that look inconspicuous to the human eye and look natural as new fashion clothing:
\begin{equation}
\mathrm{argmin}_{\tilde{X}}\sum_{k=1}^{K}(D_1(\tilde{P}_k, P_k) + D_2(\tilde{P}_k, S)).
\label{eq:robustness}
\end{equation}

For \textit{robustness}, FashionAdv uses the expectation over transformation (EOT) framework to formalize the adversarial attack problem~\cite{Athalye-2018-ICML}. In particular, it optimizes the adversarial image $\tilde{X}$ to minimize the outputs $\Psi_t(\tilde{P}_k)$ of the attacked person segmentation network (\ie, classification score and segmentation mask) in expectation:
\begin{equation}
\mathrm{argmin}_{\tilde{X}}\mathbb{E}_{t\sim T}\sum_{k=1}^{K}\Psi_t(\tilde{P}_k),
\label{eq:naturalness}
\end{equation}
where $t$ denotes a perturbation transformation, $T$ is the distribution over all possible transformations $\{t\}$, and $\Psi_t(\cdot)$ are the outputs of the attacked person segmentation network (\ie, classification score and segmentation mask) over the transformed image.

\subsection{Fashion-Guided Adversarial Attack Framework}

Figure \ref{fig:FashionAdv} depicts the FashionAdv workflow. Given an original target image, FashionAdv performs four basic steps: 1) segment the clothing regions to identify attackable regions in the image, 2) search for and select a fashion style image in a fashion corpus for use in generating an adversarial texture, 3) use the adversarial texture to create an adversarial image that is then optimized for use in an adversarial attack, and 4) use the image to spoof a person segmentation network.

FashionAdv first identifies regions in the target image in which the textures can be modified within an arbitrary range and still look natural to the human eye. The effect on image quality is minimized by focusing on only the clothing regions. These regions are segmented by using the Self-Correction for Human Parsing (SCHP)~\cite{Li-2019}, which exhibited superior performance on the Look into Person (LIP) challenge~\cite{Gong-CVPR2017}. We used the off-the-shelf model pre-trained on the LIP dataset~\cite{Gong-CVPR2017}, which was provided by the authors. FashionAdv combines semantic maps of the "upper-clothes, dress, coat, pants, skirt, and jumpsuits" labels, outputting a binary mask of the clothing regions. This mask is used to protect unattackable regions in the target image.

Next, FashionAdv automatically selects a fashion style image from a fashion style corpus we assembled. To improve the naturalness of the target image’s texture, the fashion style image closest in texture to the target image is selected by minimizing the texture transfer cost in the feature space for the target image~\cite{Bhattad-ICLR2020}. We assembled the fashion style corpus by manually selecting different raw images in the DeepFashion2 dataset~\cite{Yuying-CVPR2019}. We initially selected images of only clothing, without a person in the image. Then, we manually discarded images with patterns or materials similar to those of other images. We ended up with 140 style images in our corpus, each with a distinct style. 

FashionAdv then initializes adversarial noise and overlays it on the mask-covered target image, outputting an adversarial texture. Using the adversarial texture and the fashion style image, FashionAdv performs robustness training (\cf Section~\ref{sec:training}) to optimize the adversarial texture directly. In this training process, the adversarial texture is gradually modified to minimize both the classification score and segmentation mask produced by the attacked network while maintaining the texture’s naturalness. The adversarial image following mask removal should be robust against various transformations in cyberspace. Finally, the optimized adversarial image is directly fed into the person segmentation network to be attacked. 

\subsection{Loss Function}
\label{sec:loss}

The total loss is a combination of adversarial loss $\mathcal{L}_{adv}$ and naturalness loss $\mathcal{L}_{nat}$ in the adversarial image:
\begin{equation}
\mathcal{L} = \alpha\mathcal{L}_{adv} + \mathcal{L}_{nat}.
\label{eq:loss}
\end{equation}

Adversarial loss $\mathcal{L}_{adv}$ is directly obtained from the target person segmentation network (\ie, YOLACT) and is defined as the combination of classification loss $\mathcal{L}_{cls}$ (\ie, softmax loss) and segmentation loss $\mathcal{L}_{mask}$ (\ie, pixel-wise binary cross entropy). This is described in detail elsewhere~\cite{Bolya-ICCV2019}.

Naturalness loss $\mathcal{L}_{nat}$ as introduced here consists of inconspicuous loss $\mathcal{L}_{inc}$ and texture transfer loss $\mathcal{L}_{tex}$. Inconspicuous loss $\mathcal{L}_{inc}$ helps make the adversarial texture inconspicuous to the human eye while texture loss $\mathcal{L}_{tex}$ helps create a new fashion texture for clothing.
\begin{equation}
\mathcal{L}_{nat} = \mathcal{L}_{inc} + \beta\mathcal{L}_{tex},
\end{equation}

Inconspicuous loss is used to make the adversarial image $\tilde{X}$ similar to the original image $X$ so that the changes are inconspicuous to the human eye: 
\begin{equation}
\mathcal{L}_{inc} = \lambda_1\mathcal{L}_{sim}(X, \tilde{X}) + \lambda_2\mathcal{L}_{tv}(\tilde{X}),
\end{equation}
where $\mathcal{L}_{sim}$ is computed as the multi-scale structural similarity index measure (MS-SSIM) between $\tilde{X}$ and $X$~\cite{Wang-ACSSC2003}, and $\mathcal{L}_{tv}$ stands for the total variation loss, which is helpful for reducing noise in the adversarial texture~\cite{Mahendran-CVPR2015}.

Inspired by the efficient performance of style transfer~\cite{Bhattad-ICLR2020, Duan-CVPR2020}, we use texture loss to generate a new clothing texture with large perturbations that still looks natural to the human eye. It consists of content loss $\mathcal{L}_{c}$, which helps preserve the content of the original image $X$, and style loss $\mathcal{L}_{s}$, which helps to generate a new style from the fashion style image $S$:
\begin{equation}
\mathcal{L}_{tex} = \sum_{l \in L_c}w_l\mathcal{L}_{c}^{l}(\tilde{X},X) + \sum_{l\in L_s}w_l\mathcal{L}_{s}^{l}(\tilde{X},S),
\end{equation}
where $l$ is the $l$-th feature (\eg, the $l$-th layer of the network) in the sets of content layers $L_{c}$ and style layers $L_{s}$, and $w_l=1/C_{l+1}^2$ indicates normalized weights used to configure the layer preferences, with $C_l$ being the number of filter maps in layer $l$.

The content of the adversarial image $\tilde{X}$ may appear different from that of the original image $X$. The content loss is used to ensure that the adversarial image can preserve the content of the original image in the deep representation space through mean squared error minimization:
 \begin{equation}
\mathcal{L}_{c}^{l}(\tilde{X},X) = \left | \left | \phi_l(\tilde{X}) - \phi_l(X) \right | \right |_2^2,
\end{equation}
where $\phi_l(\cdot)$ denotes the feature extracted from the $l$-th layer in a set of content layers $L_c$ of feature extractor $\phi$.

The style of the fashion style image is transferred to the adversarial image by using the style loss, which is defined by the difference in their style representations:
\begin{equation}
\mathcal{L}_{s}^{l}(\tilde{X},S) = \frac{\left | \left | \mathcal{G}(\phi_l(\tilde{X}), \phi_{l+1}(\tilde{X})) - \mathcal{G}(\phi_l(S), \phi_{l+1}(S)) \right | \right |_2^2}{std(\mathcal{G}(\phi_l(\tilde{X}), \phi_{l+1}(\tilde{X})))},
\end{equation}
where $\mathcal{G}$ is the Gram matrix of features extracted from a set of style layers $L_s$ of feature extractor $\phi$.

\subsection{Robustness Training}
\label{sec:training}

The performance of a person segmentation network is affected by fluctuations such as a shift in the viewpoint and by other transformations in the real world. In early work~\cite{Athalye-2018-ICML}, the EOT framework was used to conduct adversarial attacks on classifiers by adding random distortions in the optimization to make the perturbations more robust. In this work, we adapt the EOT framework to train a robust adversarial attack for use against a target person segmentation network. The training was done by simulating realistic situations with a series of random image transformations in cyberspace. We included camera-related transformations for two reasons: (1) they have been proven to be effective in making adversarial perturbations robust~\cite{Athalye-2018-ICML}, and (2) they make our method extendable to the physical domain. From Eqs.~(\ref{eq:robustness}), (\ref{eq:naturalness}), and (\ref{eq:loss}), our EOT-based training is defined as: 
\begin{equation}
\mathrm{argmin}_{\tilde{X}}\mathbb{E}_{t\sim T}\alpha\mathcal{L}_{adv} + \mathcal{L}_{nat}.
\label{eq:eot}
\end{equation}

\begin{figure}[t!]
	\centering
	\includegraphics[width=1\linewidth]{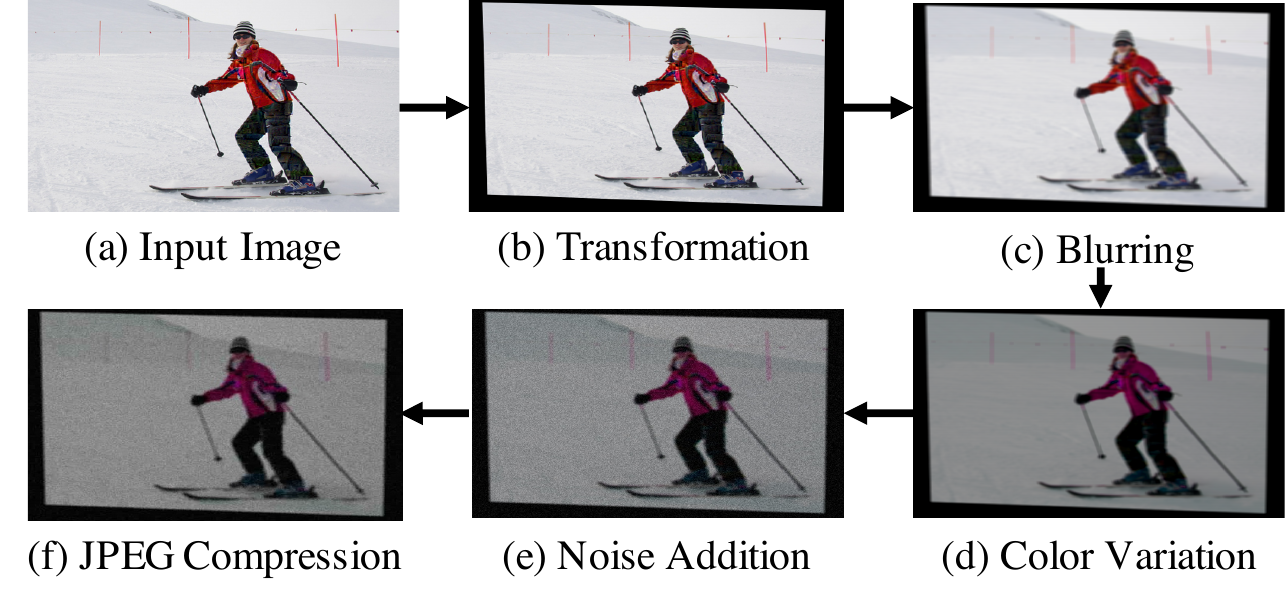}
	\caption{Image perturbation pipeline.}
	\label{fig:perturbation}
\end{figure}

Minimizing Eq.~\ref{eq:eot} results in the optimized adversarial image substantially degrading the performance of the detector, thereby resulting in a high spoofing rate. To improve the generation of adversarial images under various conditions, we used a series of perturbation transformations to simulate the cyberspace condition fluctuations. Figure \ref{fig:perturbation} depicts the image perturbation pipeline.

First, we simulated \textbf{perspective transformations} of different camera views by generating a random homography matrix in the range $[-0.2, 0.2]$ to transform the original image. Next, we simulated \textbf{image blurring} by using a Gaussian blur with kernel size and standard deviation randomly sampled in $\{1, 3, 5, 7, 9, 11\}$ and $[0.1, 3]$, respectively. We then approximated \textbf{color variation} by shifting the color values of each channel in terms of hue, saturation, brightness, and contrast in the range $[-0.02, 0.02]$ for hue and $[-0.2, 0.2]$ for the others. We also took into account \textbf{image noise} by using a uniform noise model in the range $[-0.02, 0.02]$. Finally, we applied an \textbf{approximation of JPEG compression}~\cite{Shin-NIPSW2017} with a quality factor (QF) generated uniformly in the range $[18, 22]$. We note that the original image color values were normalized in the range $[0, 1]$, and then the color values after transformations were clipped to $[0, 1]$.

\subsection{Implementation Details}

This section describes the implementation details of our approach. We implemented FashionAdv in PyTorch and conducted experiments on a computer with 32-GB RAM and a Tesla P100 GPU. We employed the ImageNet pre-trained VGG-19 network~\cite{Simonyan-ILSVRC2014} as the feature extractor. We used $L_c=\{\textit{conv4\_2}\}$ to present the content, and $L_s=\{\textit{conv1\_1, conv2\_1, conv3\_1, conv4\_1, conv5\_1}\}$ to present the style, similar to previous usage~\cite{Bhattad-ICLR2020}.

We set the weights of the loss function empirically: $\alpha=0.2, \beta=5*10^3, \lambda_1=0.75,$ and $\lambda_2=10^{-6}$. Each target image was trained using the EOT framework for 200 iterations, with the Adam optimization~\cite{Diederik-ICLR2015} and a learning rate of 0.02.

As our attack target, we selected the YOLACT, one of the first networks aimed at real-time instance segmentation. We used the pre-trained model on the MS-COCO dataset~\cite{Lin-ECCV2014}, which was provided by the authors. 

\begin{table*}[t!]
\centering
\small
\caption{Comparison of performance between FashionAdv and conventional methods in terms of robustness as represented by average precision (AP) against JPEG compression for various quality factors (QFs). The two best results are shown in \textcolor{blue}{\textbf{blue}} and \textcolor{red}{\textbf{red}}, respectively. FashionAdv significantly outperformed the conventional methods.}
\label{tab:sota}
\begin{tabular}{l|c|c|c|c|c|c|c|c}
\hline
\multirow{2}{*}{\textbf{Method}} & \multirow{2}{*}{\textbf{Conference/Year}} & \multirow{2}{*}{\textbf{No}} & \multicolumn{6}{c}{\textbf{JPEG Compression}} \\ 
\cline{4-9} 
& & \textbf{Compression} & \textbf{QF 100} & \textbf{QF 80} & \textbf{QF 60} & \textbf{QF 40} & \textbf{QF 20} & \textbf{QF 10} \\ 
\hline
Random Noise & - & 58.37 & 58.36 & 57.89 & 57.62 & 56.96 & 54.32 & 40.65 \\ 
FGSM~\cite{Goodfellow-2014-ICLR} & ICLR 2014 & 49.71 & 49.93 & 51.33 & 52.63 & 52.90 & 51.97 & 43.30 \\ 
BIM~\cite{Kurakin-2016-ICLRW} & ICLRW 2016 & 40.87 & 41.63 & 50.11 & 54.82 & 58.13 & 59.16 & 46.65 \\ 
DeepFool~\cite{Moosavi-CVPR2016} & CVPR 2016 & 42.23 & 49.99 & 63.33 & 64.00 & 63.34 & 60.00 & 46.35 \\ 
PGD~\cite{Madry-ICLR2018} & ICLR 2018 & \textcolor{blue}{3.74} & \textcolor{blue}{4.70} & 56.05 & 61.12 & 61.66 & 45.65 & 45.65 \\ 
ColorFool~\cite{Shamsabadi-2020-CVPR} & CVPR 2020 & 46.14 & 48.66 & 50.65 & 50.92 & 50.41 & 47.27 & 33.77 \\ 
cAdv~\cite{Bhattad-ICLR2020} & ICLR 2020 & 26.16 & 29.70 & 44.74 & 48.05 & 49.24 & 46.66 & 33.14 \\ 
tAdv~\cite{Bhattad-ICLR2020} & ICLR 2020 & 29.05 & 30.77 & 49.62 & 55.27 & 58.07 & 56.98 & 45.44 \\ 
\rowcolor{lightgray} \textbf{FashionAdv (rnd)} & \textbf{CVPRW 2021} & \textbf{21.30} & \textbf{21.83} & \textbf{\textcolor{red}{24.36}} & \textbf{\textcolor{red}{25.70}} &  \textbf{\textcolor{red}{26.05}} & \textbf{\textcolor{red}{25.63}} & \textbf{\textcolor{red}{22.20}} \\
\rowcolor{lightgray} \textbf{FashionAdv} & \textbf{CVPRW 2021} & \textcolor{red}{\textbf{18.40}} & \textcolor{red}{\textbf{18.52}} &  \textcolor{blue}{\textbf{19.90}} & \textcolor{blue}{\textbf{20.36}} &  \textcolor{blue}{\textbf{20.49}} & \textcolor{blue}{\textbf{20.59}} &  \textcolor{blue}{\textbf{20.82}}  \\ 
\hline
\end{tabular}
\vspace{-2mm}
\end{table*}


\section{Experiments}
\label{sec:results}

\subsection{Benchmark Dataset and Evaluation Criteria}

To evaluate the proposed method, we selected images containing at least one person from the MS-COCO dataset~\cite{Lin-ECCV2014} and ended up with 1000 images. We evaluated robustness using Average Precision (AP)~\cite{Lin-ECCV2014} and naturalness using Structural Similarity Index Measure (SSIM)~\cite{Wang-TIP2004}. To evaluate the effectiveness of methods, we computed AP by comparing the results of person segmentation before and after an attack.

\subsection{Comparison with State-of-the-Arts}

We compared the performance of FashionAdv with those of state-of-the-art adversarial example methods: FGSM~\cite{Goodfellow-2014-ICLR}, BIM~\cite{Kurakin-2016-ICLRW}, PGD~\cite{Madry-ICLR2018}, DeepFool~\cite{Moosavi-CVPR2016}, ColorFool~\cite{Shamsabadi-2020-CVPR}, cAdv~\cite{Bhattad-ICLR2020}, and tAdv~\cite{Bhattad-ICLR2020}. These methods were originally developed to attack classifiers; we thus adapted them to attack the classification sub-network in the person segmentation network (\ie, YOLACT). Instance segmentation networks are harder to fool than classifiers~\cite{Wu-ECCV2020}, and our empirical experiments revealed that the conventional methods with their default parameter settings are even weaker than a random noise attack, which is supposedly the weakest type of attack. We thus determined that it was pointless to use the default parameter settings to attack person segmentation networks. Therefore, we adjusted their parameter settings by using a gird search to make the compared methods more reliable than random noise methods in terms of AP score. The parameter settings and source codes are publicly released on our project page.

\begin{figure}[t!]
	\centering
	\includegraphics[width=1\linewidth]{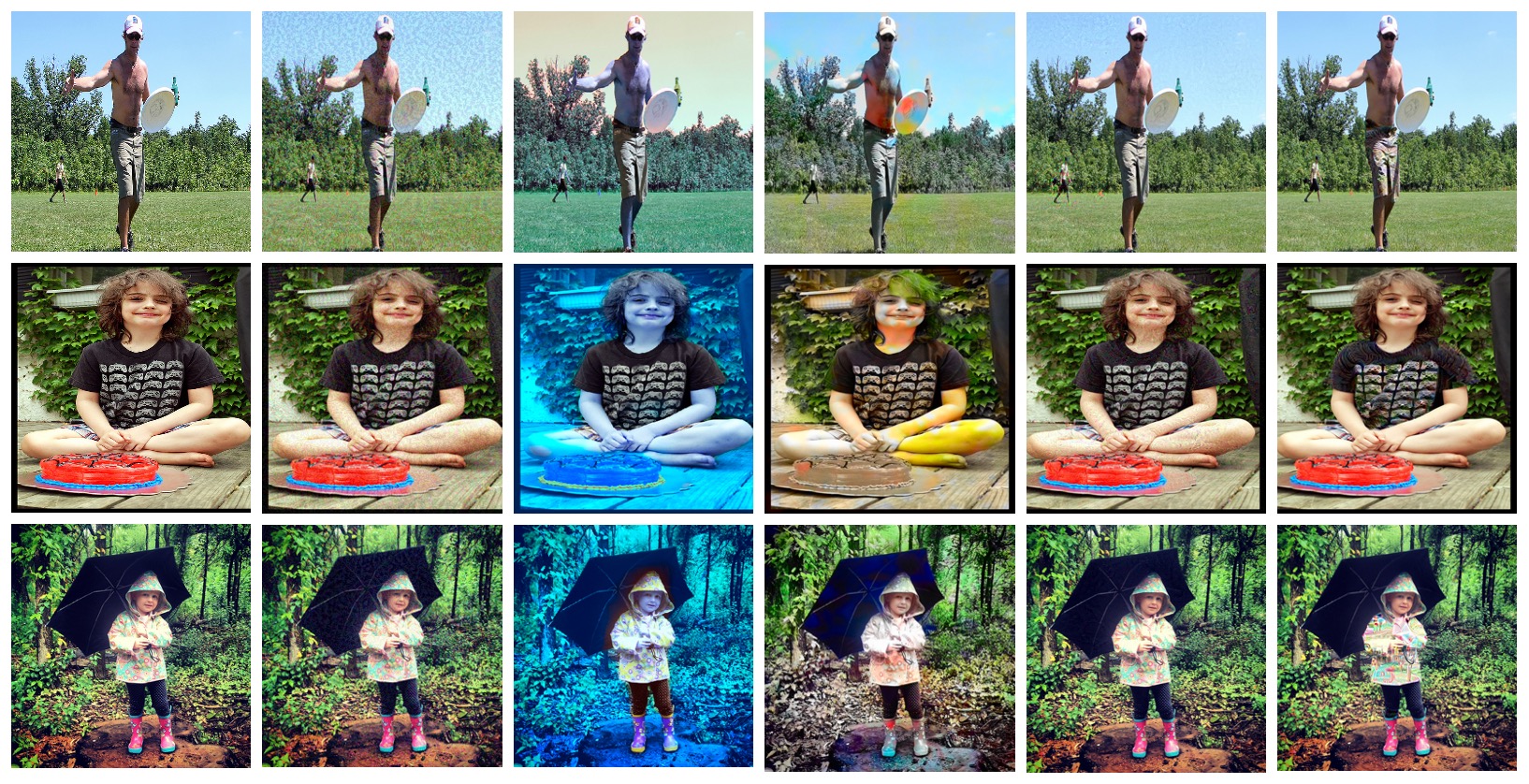}
	\caption{Comparison of the resulting images of different methods. From left to right: the original image, FGSM~\cite{Goodfellow-2014-ICLR}, ColorFool~\cite{Shamsabadi-2020-CVPR}, cAdv~\cite{Bhattad-ICLR2020}, tAdv~\cite{Bhattad-ICLR2020}, and our FashionAdv. Changes in image with FashionAdv are inconspicuous. (Best viewed online in color with zoom-in.)}
	\label{fig:visualization}
\end{figure}

Figure~\ref{fig:trade_off} illustrates the trade-off between naturalness (\ie, SSIM) and robustness (\ie, AP) for the compared methods. FashionAdv significantly outperformed the other methods on robustness against JPEG compression with different QFs (\ie, 10, 40, and 80): the AP scores of FashionAdv were around 20 while those of the other methods were higher than 30. FashionAdv also achieved competitive naturalness and was a top method in achieving the highest SSIM scores. As shown in Fig.~\ref{fig:visualization}, the adversarial images generated by FashionAdv looked the most natural.

Table~\ref{tab:sota} compares the results for robustness in detail. FashionAdv (rnd) is our proposed method using fashion style random selection strategy (see section~\ref{sec:ablation} for more detail). Our method significantly outperformed state-of-the-art methods and achieved the highest robustness against image compression. The conventional methods were not robust even for uncompressed images (except for PGD) and JPEG compressed images. Their AP scores were higher than 30 and rapidly increased for the JPEG compressed images. In contrast, the scores for FashionAdv were lower than 20 on uncompressed images and remained stable at around 20 for the JPEG compressed images. 

\begin{table}[t!]
\centering
\small
\caption{Contributions of loss components to final result.}
\label{tab:ablations_loss}
\begin{tabular}{cccc|c|c}
\hline
\textbf{$\mathcal{L}_{adv}$} & \textbf{$\mathcal{L}_{tex}$} &
\textbf{$\mathcal{L}_{sim}$} & \textbf{$\mathcal{L}_{tv}$} & \textbf{AP} & \textbf{SSIM} \\ 
\hline
\checkmark &  &  & & 18.68  & 0.954 \\
\checkmark & \checkmark &  &  & 18.62 & 0.954 \\
\checkmark & \checkmark & \checkmark &  & 18.60 & 0.955 \\
\checkmark & \checkmark &  & \checkmark & 18.63 & 0.956 \\
\checkmark & \checkmark & \checkmark & \checkmark & \textbf{18.40} & \textbf{0.958} \\ 
\hline
\end{tabular}
\end{table}

\subsection{Ablation Study}
\label{sec:ablation}
We investigated the effect of different components of our proposed FashionAdv, such as targeted attack, components of loss function, fashion style selection, and robustness analysis.

\textbf{Importance of Attacked Targets. } Previous attack methods target only the classification module in deep networks, so they work only on image classification. For person segmentation, both the classification and segmentation modules must be targeted. Indeed, an experiment we conducted on attacked targets revealed that targeting both the classification and segmentation modules significantly reduced AP by 18.35, from 36.75 (corresponding to target only classification module by setting $\mathcal{L}_{mask}=0$) to 18.40.

\textbf{Contributions of Loss Components. } Table~\ref{tab:ablations_loss} shows the contributions of the loss components to the final result. $\mathcal{L}_{tex}$ helps to generate strong adversarial textures while $\mathcal{L}_{sim}$ and $\mathcal{L}_{tv}$ improve the naturalness and inconspicuousness of the generated textures. Integrating these loss components is thus useful in synthesizing robust and natural adversarial textures.

\begin{figure}[t!]
	\centering
	\includegraphics[width=1\linewidth]{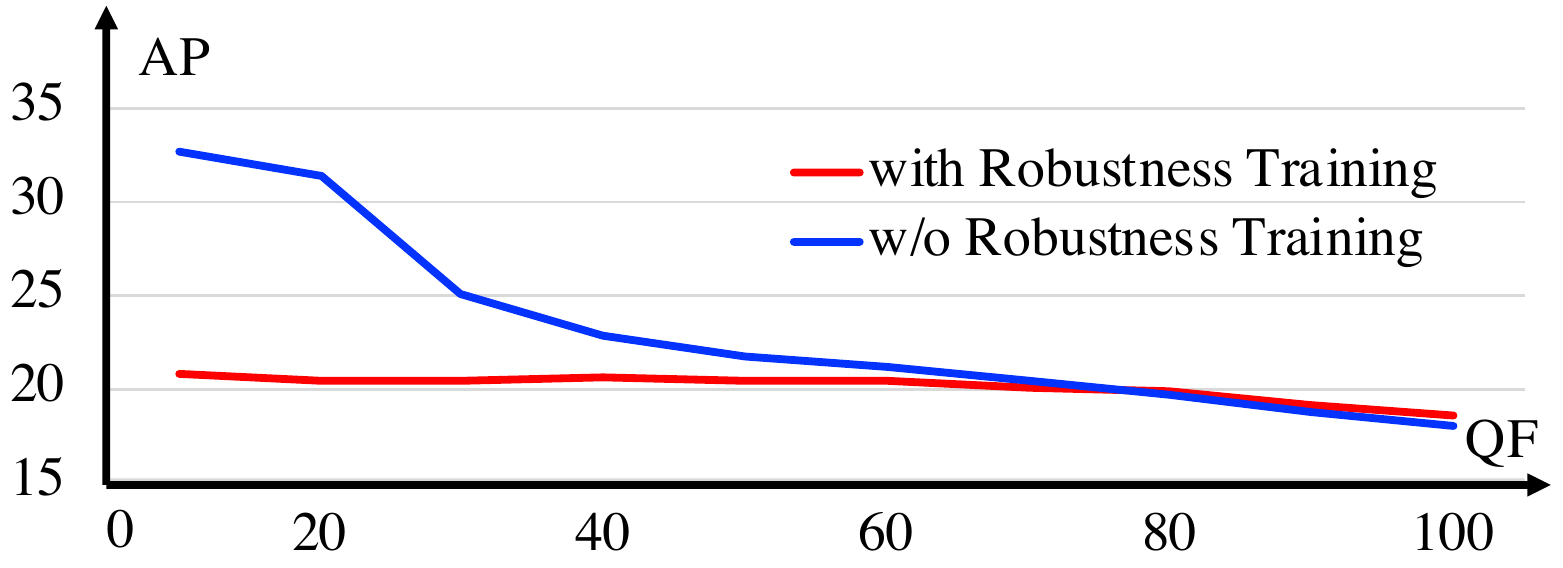}
	\caption{Robustness against JPEG compression for different quality factors QFs.}
	\label{fig:jpeg_compression}
	\vspace{-3mm}
\end{figure}

\textbf{Fashion Style Selection. } We employed two strategies for fashion style selection. In the first one, a fashion style image is randomly selected from our style corpus for each target image, denoted by \textit{FashionAdv (rnd)}. In the second one, denoted by FashionAdv, fashion style image that minimized the texture transfer cost in the feature space is selected for each target image, similar to Bhattad et al.~\cite{Bhattad-ICLR2020}. 
Using the optimal selection strategy effectively reduced AP by 3 to 5 compared with using the random selection strategy (\cf~Table~\ref{tab:sota}). With this optimized strategy, the fashion style image closest to the target image is selected, resulting in more natural adversarial examples. This result also indicates the crucial role of using guided-fashion style images for synthesizing adversarial textures.


\textbf{Robustness Against Image Storage. } To evaluate the robustness of FashionAdv against image storage, \ie, against compression and saving in JPEG format, we investigated using JPEG cues in the robustness training. We compared training with and without JPEG cue for different JPEG quality factors (QFs). As shown in Fig.~\ref{fig:jpeg_compression}, training adversarial examples with JPEG cues kept the AP score around 20 for all QFs. In contrast, the AP score of training without JPEG cues increased to 33 at small QFs. Hence, integrating JPEG cues into the robustness training effectively defends against JPEG compression for various QFs. 

\textbf{Robustness Against Image Manipulations. } To further exhibit the robustness of FashionAdv against image manipulations, we investigated the performance of person segmentation (in terms of AP) on several image editing operations: scaling, blurring, color jitter, and noise addition. We conducted manipulation attacks in two modes, easy and hard. In easy mode, we used randomly scaled images (0.8 to 1.2 of original size), Gaussian blur with a $5 \times 5$ kernel, histogram equalization for color jitter, and uniform noise of 0.05. In hard mode, we used randomly scaled images (0.5 to 1.5 of original size), Gaussian blur with a $9 \times 9$ kernel, randomly changing values of each color channel in the range [-0.2, 0.2], and uniform noise of 0.15. We remark that image color values were normalized and clipped to $[0, 1]$.

The effectiveness of the robustness training, which included perspective transform, image blurring, color variation, and image noise (Section~\ref{sec:training}) was demonstrated for each attack mode. Table~\ref{tab:ablations_data_editing} shows that the robustness training significantly outperformed the standard training (\ie, non-robustness training) for both attack modes, reducing the AP scores by 30. Utilizing our image perturbation pipeline in the robustness training process was effective against all image editing operations.

Table~\ref{tab:ablations_data_editing} also shows that the completed FashionAdv (with robustness training) was significantly robust against color jitter and noise addition even in hard mode, in which the AP scores increased by less than 5. We remark that the AP for the original images was 18.04 (\cf Table~\ref{tab:sota}). FashionAdv also well defended against two more difficult image editing operations (\ie, scaling and blurring). Image scaling eliminates small details (including generated adversarial noise) for downscaled images or highlights adversarial noise for upscaled images. Image blurring can wash away adversarial noise in the images. For these two image editing operators, AP was smaller than 24, indicating that FashionAdv also well defends against these difficult attacks.

\begin{table}[t!]
\centering
\small
\caption{Robustness against image manipulations in terms of AP score.}
\label{tab:ablations_data_editing}
\resizebox{1\linewidth}{!}{
\begin{tabular}{l|c|c|c|c|c}
\hline
\begin{tabular}[l]{@{}l@{}}\textbf{Attack} \\ \textbf{Mode}\end{tabular} & \begin{tabular}[c]{@{}c@{}}\textbf{Robustness} \\ \textbf{Training}\end{tabular}  & \textbf{Scaling} & \textbf{Blurring} & \textbf{Color Jitter} & \begin{tabular}[c]{@{}c@{}}\textbf{Noise} \\ \textbf{Addition}\end{tabular} \\ 
\hline
\multirow{2}{*}{Easy} & $\checkmark$ & 19.93 & 20.75 & 19.12 & 18.04 \\ 
& $\times$ & 44.94 & 50.45 & 21.88 & 23.72 \\ 
\hline
\multirow{2}{*}{Hard} & $\checkmark$ & 23.50 & 23.66 & 22.93 & 19.23 \\ 
& $\times$ & 48.10 & 51.86 & 29.35 & 49.06 \\
\hline
\end{tabular}
}
\vspace{-3mm}
\end{table}


\section{Conclusion}
\label{sec:conclusion}

Our proposed adversarial example based method is aimed at spoofing person segmentation networks. 
FashionAdv automatically detects and attacks the clothing regions by synthesizing robust yet natural adversarial textures from guided-fashion style images, resulting in all persons in the image being invisible to the networks. Extensive testing showed that FashionAdv is not only robust against digital image filtering and compression but also produces images that appear natural to the human eye. 

FashionAdv relies on SCHP for generating binary masks of the clothing regions. If the clothing regions cannot be segmented, the generated adversarial images will be unable to spoof person segmentation networks. If the binary masks are over-segmented, manipulations can be easily perceived. Future work includes investigating better clothing segmentation approaches. It also includes further exploration of adversarial attacks on general instance segmentation.


\textbf{Acknowledgements. } This work was partially supported by JSPS KAKENHI Grants JP16H06302, JP18H04120, JP21H04907, JP20K23355, and JP21K18023, and by JST CREST Grants JPMJCR18A6 and JPMJCR20D3, including the AIP challenge program, Japan.

{\small
\bibliographystyle{ieee_fullname}
\bibliography{short_bibtex}
}

\end{document}